# On Smoothing and Inference for Topic Models

**Arthur Asuncion, Max Welling, Padhraic Smyth**
Department of Computer Science
University of California, Irvine
Irvine, CA, USA
{asuncion,welling,smyth}@ics.uci.edu

**Yee Whye Teh**
Gatsby Computational Neuroscience Unit
University College London
London, UK
ywteh@gatsby.ucl.ac.uk

## Abstract

Latent Dirichlet analysis, or topic modeling, is a flexible latent variable framework for modeling high-dimensional sparse count data. Various learning algorithms have been developed in recent years, including collapsed Gibbs sampling, variational inference, and maximum a posteriori estimation, and this variety motivates the need for careful empirical comparisons. In this paper, we highlight the close connections between these approaches. We find that the main differences are attributable to the amount of smoothing applied to the counts. When the hyperparameters are optimized, the differences in performance among the algorithms diminish significantly. The ability of these algorithms to achieve solutions of comparable accuracy gives us the freedom to select computationally efficient approaches. Using the insights gained from this comparative study, we show how accurate topic models can be learned in several seconds on text corpora with thousands of documents.

## 1　INTRODUCTION

Latent Dirichlet Allocation (LDA) [Blei et al., 2003] and Probabilistic Latent Semantic Analysis (PLSA) [Hofmann, 2001] are well-known latent variable models for high dimensional count data, such as text data in the bag-of-words representation or images represented through feature counts. Various inference techniques have been proposed, including collapsed Gibbs sampling (CGS) [Griffiths and Steyvers, 2004], variational Bayesian inference (VB) [Blei et al., 2003], collapsed variational Bayesian inference (CVB) [Teh et al., 2007], maximum likelihood estimation (ML) [Hofmann, 2001], and maximum a posteriori estimation (MAP) [Chien and Wu, 2008].

Among these algorithms, substantial performance differences have been observed in practice. For instance, Blei et al. [2003] have shown that the VB algorithm for LDA outperforms ML estimation for PLSA. Furthermore, Teh et al. [2007] have found that CVB is significantly more accurate than VB. But can these differences in performance really be attributed to the type of inference algorithm?

In this paper, we provide convincing empirical evidence that points in a different direction, namely that the claimed differences can be explained away by the different settings of two smoothing parameters (or hyperparameters). In fact, our empirical results suggest that these inference algorithms have relatively similar predictive performance when the hyperparameters for each method are selected in an optimal fashion. With hindsight, this phenomenon should not surprise us. Topic models operate in extremely high dimensional spaces (with typically more than 10,000 dimensions) and, as a consequence, the "curse of dimensionality" is lurking around the corner; thus, hyperparameter settings have the potential to significantly affect the results.

We show that the potential perplexity gains by careful treatment of hyperparameters are on the order of (if not greater than) the differences between different inference algorithms. These results caution against using generic hyperparameter settings when comparing results across algorithms. This in turn raises the question as to whether newly introduced models and approximate inference algorithms have real merit, or whether the observed difference in predictive performance is attributable to suboptimal settings of hyperparameters for the algorithms being compared.

In performing this study, we discovered that an algorithm which suggests itself in thinking about inference algorithms in a unified way – but was never proposed by itself before – performs best, albeit marginally so. More importantly, it happens to be the most computationally efficient algorithm as well.

In the following section, we highlight the similarities between each of the algorithms. We then discuss the importance of hyperparameter settings. We show accuracy results, using perplexity and precision/recall metrics, for each algorithm over various text data sets. We then focus on computational efficiency and provide timing results across algorithms. Finally, we discuss related work and conclude with future directions.

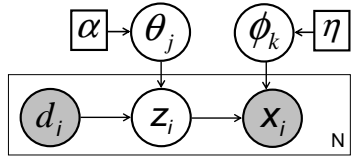

Figure 1: Graphical model for Latent Dirichlet Allocation. Boxes denote parameters, and shaded/unshaded circles denote observed/hidden variables.

## 2 INFERENCE TECHNIQUES FOR LDA

LDA has roots in earlier statistical decomposition techniques, such as Latent Semantic Analysis (LSA) [Deerwester et al., 1990] and Probabilistic Latent Semantic Analysis (PLSA) [Hofmann, 2001]. Proposed as a generalization of PLSA, LDA was cast within the generative Bayesian framework to avoid some of the overfitting issues that were observed with PLSA [Blei et al., 2003]. A review of the similarities between LSA, PLSA, LDA, and other models can be found in Buntine and Jakulin [2006].

We describe the LDA model and begin with general notation. LDA assumes the standard bag-of-words representation, where $D$ documents are each represented as a vector of counts with $W$ components, where $W$ is the number of words in the vocabulary. Each document $j$ in the corpus is modeled as a mixture over $K$ topics, and each topic $k$ is a distribution over the vocabulary of $W$ words. Each topic, $\phi_{\cdot k}$, is drawn from a Dirichlet with parameter $\eta$, while each document's mixture, $\theta_{\cdot j}$, is sampled from a Dirichlet with parameter $\alpha$[1]. For each token $i$ in the corpus, a topic assignment $z_i$ is sampled from $\theta_{\cdot d_i}$, and the specific word $x_i$ is drawn from $\phi_{\cdot z_i}$. The generative process is below:

$$\theta_{k,j} \sim \mathcal{D}[\alpha] \quad \phi_{w,k} \sim \mathcal{D}[\eta] \quad z_i \sim \theta_{k,d_i} \quad x_i \sim \phi_{w,z_i}.$$

In Figure 1, the graphical model for LDA is presented in a slightly unconventional fashion, as a Bayesian network where $\theta_{kj}$ and $\phi_{wk}$ are conditional probability tables and $i$ runs over all tokens in the corpus. Each token's document index $d_i$ is explicitly shown as an observed variable, in order to show LDA's correspondence to the PLSA model.

Exact inference (i.e. computing the posterior over the hidden variables) for this model is intractable [Blei et al., 2003], and so a variety of approximate algorithms have been developed. If we ignore $\alpha$ and $\eta$ and treat $\theta_{kj}$ and $\phi_{wk}$ as parameters, we obtain the PLSA model, and maximum likelihood (ML) estimation over $\theta_{kj}$ and $\phi_{wk}$ directly corresponds to PLSA's EM algorithm. Adding the hyperparameters $\alpha$ and $\eta$ back in leads to MAP estimation.

[1] We use symmetric Dirichlet priors for simplicity in this paper.

Treating $\theta_{kj}$ and $\phi_{wk}$ as hidden variables and factorizing the posterior distribution leads to the VB algorithm, while collapsing $\theta_{kj}$ and $\phi_{wk}$ (i.e. marginalizing over these variables) leads to the CVB and CGS algorithms. In the following subsections, we provide details for each approach.

### 2.1 ML ESTIMATION

The PLSA algorithm described in Hofmann [2001] can be understood as an expectation maximization algorithm for the model depicted in Figure 1. We start the derivation by writing the log-likelihood as,

$$\ell = \sum_i \log \sum_{z_i} P(x_i|z_i, \phi) \, P(z_i|d_i, \theta)$$

from which we derive via a standard EM derivation the updates (where we have left out explicit normalizations):

$$P(z_i|x_i, d_i) \propto P(x_i|z_i, \phi) \, P(z_i|d_i, \theta) \quad (1)$$

$$\phi_{w,k} \propto \sum_i \mathbb{I}[x_i = w, z_i = k] P(z_i|x_i, d_i) \quad (2)$$

$$\theta_{k,j} \propto \sum_i \mathbb{I}[z_i = k, d_i = j] P(z_i|x_i, d_i). \quad (3)$$

These updates can be rewritten by defining $\gamma_{wjk} = P(z = k|x = w, d = j)$, $N_{wj}$ the number of observations for word type $w$ in document $j$, $N_{wk} = \sum_j N_{wj}\gamma_{wjk}$, $N_{kj} = \sum_w N_{wj}\gamma_{wjk}$, $N_k = \sum_w N_{wk}$ and $N_j = \sum_k N_{kj}$,

$$\phi_{w,k} \leftarrow N_{wk}/N_k \qquad \theta_{k,j} \leftarrow N_{kj}/N_j \, .$$

Plugging these expressions back into the expression for the posterior (1) we arrive at the update,

$$\gamma_{wjk} \propto \frac{N_{wk} \, N_{kj}}{N_k} \quad (4)$$

where the constant $N_j$ is absorbed into the normalization. Hofmann [2001] regularizes the PLSA updates by raising the right hand side of (4) to a power $\beta > 0$ and searching for the best value of $\beta$ on a validation set.

### 2.2 MAP ESTIMATION

We treat $\phi, \theta$ as random variables from now on. We add Dirichlet priors with strengths $\eta$ for $\phi$ and $\alpha$ for $\theta$ respectively. This extension was introduced as "latent Dirichlet allocation" in Blei et al. [2003].

It is possible to optimize for the MAP estimate of $\phi, \theta$. The derivation is very similar to the ML derivation in the previous section, except that we now have terms corresponding to the log of the Dirichlet prior which are equal to $\sum_{wk}(\eta - 1)\log \phi_{wk}$ and $\sum_{kj}(\alpha - 1)\log \theta_{kj}$. After working through the math, we derive the following update (de Freitas and Barnard [2001], Chien and Wu [2008]),

$$\gamma_{wjk} \propto \frac{(N_{wk} + \eta - 1)(N_{kj} + \alpha - 1)}{(N_k + W\eta - W)} \quad (5)$$

where $\alpha, \eta > 1$. Upon convergence, MAP estimates are obtained:

$$\hat{\phi}_{wk} = \frac{N_{wk} + \eta - 1}{N_k + W\eta - W} \quad \hat{\theta}_{kj} = \frac{N_{kj} + \alpha - 1}{N_j + K\alpha - K}. \quad (6)$$

## 2.3 VARIATIONAL BAYES

The variational Bayesian approximation (VB) to LDA follows the standard variational EM framework [Attias, 2000, Ghahramani and Beal, 2000]. We introduce a factorized (and hence approximate) variational posterior distribution:

$$Q(\phi, \theta, \mathbf{z}) = \prod_k q(\phi_{\cdot,k}) \prod_j q(\theta_{\cdot,j}) \prod_i q(z_i).$$

Using this assumption in the variational formulation of the EM algorithm [Neal and Hinton, 1998] we readily derive the VB updates analogous to the ML updates of Eqns. 2, 3 and 1:

$$q(\phi_{\cdot,k}) = \mathcal{D}[\eta + N_{\cdot,k}], \quad N_{wk} = \sum_i q(z_i = k)\delta(x_i, w) \quad (7)$$

$$q(\theta_{\cdot,j}) = \mathcal{D}[\alpha + N_{\cdot,j}], \quad N_{kj} = \sum_i q(z_i = k)\delta(d_i, j) \quad (8)$$

$$q(z_i) \propto \exp\left(E[\log \phi_{x_i, z_i}]_{q(\phi)} E[\log \theta_{z_i, d_i}]_{q(\theta)}\right). \quad (9)$$

We can insert the expression for $q(\phi)$ at (7) and $q(\theta)$ at (8) into the update for $q(z)$ in (9) and use the fact that $E[\log X_i]_{\mathcal{D}(X)} = \psi(X_i) - \psi(\sum_j X_j)$ with $\psi(\cdot)$ being the "digamma" function. As a final observation, note that there is nothing in the free energy that would render any differently the distributions $q(z_i)$ for tokens that correspond to the same word-type $w$ in the same document $j$. Hence, we can simplify and update a single prototype of that equivalence class, denoted as $\gamma_{wjk} \triangleq q(z_i = k)\delta(x_i, w)\delta(d_i, j)$ as follows,

$$\gamma_{wjk} \propto \frac{\exp(\psi(N_{wk} + \eta))}{\exp(\psi(N_k + W\eta))} \exp(\psi(N_{kj} + \alpha)). \quad (10)$$

We note that $\exp(\psi(n)) \approx n - 0.5$ for $n > 1$. Since $N_{wk}$, $N_{kj}$, and $N_k$ are aggregations of expected counts, we expect many of these counts to be greater than 1. Thus, the VB update can be approximated as follows,

$$\gamma_{wjk} \approx \propto \frac{(N_{wk} + \eta - 0.5)}{(N_k + W\eta - 0.5)} (N_{kj} + \alpha - 0.5) \quad (11)$$

which exposes the relation to the MAP update in (5).

In closing this section, we mention that the original VB algorithm derived in Blei et al. [2003] was a hybrid version between what we call VB and ML here. Although they did estimate variational posteriors $q(\theta)$, the $\phi$ were treated as parameters and were estimated through ML.

## 2.4 COLLAPSED VARIATIONAL BAYES

It is possible to marginalize out the random variables $\theta_{kj}$ and $\phi_{wk}$ from the joint probability distribution. Following a variational treatment, we can introduce variational posteriors over $z$ variables which is once again assumed to be factorized: $Q(z) = \prod_i q(z_i)$. This collapsed variational free energy represents a strictly better bound on the (negative) evidence than the original VB [Teh et al., 2007]. The derivation of the update equation for the $q(z_i)$ is slightly more complicated and involves approximations to compute intractable summations. The update is given below [2]:

$$\gamma_{ijk} \propto \frac{N_{wk}^{\neg ij} + \eta}{N_k^{\neg ij} + W\eta} \left(N_{kj}^{\neg ij} + \alpha\right) \exp\left(-\frac{V_{kj}^{\neg ij}}{2(N_{kj}^{\neg ij} + \alpha)^2}\right.$$
$$\left. - \frac{V_{wk}^{\neg ij}}{2(N_{wk}^{\neg ij} + \eta)^2} + \frac{V_k^{\neg ij}}{2(N_k^{\neg ij} + W\eta)^2}\right). \quad (12)$$

$N_{kj}^{\neg ij}$ denotes the expected number of tokens in document $j$ assigned to topic $k$ (excluding the current token), and can be calculated as follows: $N_{kj}^{\neg ij} = \sum_{i' \neq i} \gamma_{i'jk}$. For CVB, there is also a variance associated with each count: $V_{kj}^{\neg ij} = \sum_{i' \neq i} \gamma_{i'jk}(1 - \gamma_{i'jk})$. For further details we refer to Teh et al. [2007].

The update in (12) makes use of a second-order Taylor expansion as an approximation. A further approximation can be made by using only the zeroth-order information[3]:

$$\gamma_{ijk} \propto \frac{N_{wk}^{\neg ij} + \eta}{N_k^{\neg ij} + W\eta} \left(N_{kj}^{\neg ij} + \alpha\right). \quad (13)$$

We refer to this approximate algorithm as CVB0.

## 2.5 COLLAPSED GIBBS SAMPLING

MCMC techniques are available to LDA as well. In collapsed Gibbs sampling (CGS) [Griffiths and Steyvers, 2004], $\theta_{kj}$ and $\phi_{wk}$ are integrated out (as in CVB) and sampling of the topic assignments is performed sequentially in the following manner:

$$P(z_{ij} = k | z^{\neg ij}, x_{ij} = w) \propto \frac{N_{wk}^{\neg ij} + \eta}{N_k^{\neg ij} + W\eta} \left(N_{kj}^{\neg ij} + \alpha\right). \quad (14)$$

$N_{wk}$ denotes the number of word tokens of type $w$ assigned to topic $k$, $N_{kj}$ is the number of tokens in document $j$ assigned to topic $k$, and $N_k = \sum_w N_{wk}$. $N^{\neg ij}$ denotes the count with token $ij$ removed. Note that standard non-collapsed Gibbs sampling over $\phi$, $\theta$, and $z$ can also be performed, but we have observed that CGS mixes more quickly in practice.

---

[2] For convenience, we switch back to the conventional indexing scheme for LDA where $i$ runs over tokens in document $j$.

[3] The first-order information becomes zero in this case.

## 2.6 COMPARISON OF ALGORITHMS

A comparison of update equations (5), (11), (12), (13), (14) reveals the similarities between these algorithms. All of these updates consist of a product of terms featuring $N_{wk}$ and $N_{kj}$ as well as a denominator featuring $N_k$. These updates resemble the Callen equations [Teh et al., 2007], which the true posterior distribution must satisfy (with $Z$ as the normalization constant):

$$P(z_{ij} = k|x) = E_{p(z^{\neg ij}|x)}\left[\frac{1}{Z}\frac{(N_{wk}^{\neg ij} + \eta)}{(N_k^{\neg ij} + W\eta)}(N_{kj}^{\neg ij} + \alpha)\right].$$

We highlight the striking connections between the algorithms. Interestingly, the probabilities for CGS (14) and CVB0 (13) are exactly the same. The only difference is that CGS samples each topic assignment while CVB0 deterministically updates a discrete distribution over topics for each token. Another way to view this connection is to imagine that CGS can sample each topic assignment $z_{ij}$ $R$ times using (14) and maintain a distribution over these samples with which it can update the counts. As $R \to \infty$, this distribution will be exactly (13) and this algorithm will be CVB0. The fact that algorithms like CVB0 are able to propagate the entire uncertainty in the topic distribution during each update suggests that deterministic algorithms should converge more quickly than CGS.

CVB0 and CVB are almost identical as well, the distinction being the inclusion of second-order information for CVB.

The conditional distributions used in VB (11) and MAP (5) are also very similar to those used for CGS, with the main difference being the presence of offsets of up to $-0.5$ and $-1$ in the numerator terms for VB and MAP, respectively. Through the setting of hyperparameters $\alpha$ and $\eta$, these extra offsets in the numerator can be eliminated, which suggests that these algorithms can be made to perform similarly with appropriate hyperparameter settings.

Another intuition which sheds light on this phenomenon is as follows. Variational methods like VB are known to underestimate posterior variance [Wang and Titterington, 2004]. In the case of LDA this is reflected in the offset of -0.5: typical values of $\phi$ and $\theta$ in the variational posterior tend to concentrate more mass on the high probability words and topics respectively. We can counteract this by incrementing the hyperparameters by 0.5, which encourages more probability mass to be smoothed to all words and topics. Similarly, MAP offsets by -1, concentrating even more mass on high probability words and topics, and requiring even more smoothing by incrementing $\alpha$ and $\eta$ by 1.

Other subtle differences between the algorithms exist. For instance, VB subtracts only 0.5 from the denominator while MAP removes $W$, which suggests that VB applies more smoothing to the denominator. Since $N_k$ is usually large, we do not expect this difference to play a large role in learning. For the collapsed algorithms (CGS, CVB), the counts $N_{wk}, N_{kj}, N_k$ are updated after each token update. Meanwhile, the standard formulations of VB and MAP update these counts only after sweeping through all the tokens. This update schedule may affect the rate of convergence [Neal and Hinton, 1998]. Another difference is that the collapsed algorithms remove the count for the current token $ij$.

As we will see in the experimental results section, the performance differences among these algorithms that were observed in previous work can be substantially reduced when the hyperparameters are optimized for each algorithm.

## 3 THE ROLE OF HYPERPARAMETERS

The similarities between the update equations for these algorithms shed light on the important role that hyperparameters play. Since the amount of smoothing in the updates differentiates the algorithms, it is important to have good hyperparameter settings. In previous results [Teh et al., 2007, Welling et al., 2008b, Mukherjee and Blei, 2009], hyperparameters for VB were set to small values like $\alpha = 0.1$, $\eta = 0.1$, and consequently, the performance of VB was observed to be significantly suboptimal in comparison to CVB and CGS. Since VB effectively adds a discount of up to $-0.5$ in the updates, greater values for $\alpha$ and $\eta$ are necessary for VB to perform well. We discuss hyperparameter learning and the role of hyperparameters in prediction.

### 3.1 HYPERPARAMETER LEARNING

It is possible to learn the hyperparameters during training. One approach is to place Gamma priors on the hyperparameters ($\eta \sim G[a, b], \alpha \sim G[c, d]$) and use Minka's fixed-point iterations [Minka, 2000], e.g.:

$$\alpha \leftarrow \frac{c - 1 + \hat{\alpha}\sum_j \sum_k \left[\Psi(N_{kj} + \hat{\alpha}) - \Psi(\hat{\alpha})\right]}{d + K\sum_j \left[\Psi(N_j + K\hat{\alpha}) - \Psi(K\hat{\alpha})\right]}.$$

Other ways for learning hyperparameters include Newton-Raphson and other fixed-point techniques [Wallach, 2008], as well as sampling techniques [Teh et al., 2006]. Another approach is to use a validation set and to explore various settings of $\alpha, \eta$ through grid search. We explore several of these approaches later in the paper.

### 3.2 PREDICTION

Hyperparameters play a role in prediction as well. Consider the update for MAP in (5) and the estimates for $\phi_{wk}$ and $\theta_{kj}$ (6) and note that the terms used in learning are the same as those used in prediction. Essentially the same can be said for the collapsed algorithms, since the following Rao-Blackwellized estimates are used, which bear resemblance to terms in the updates (14), (13):

$$\hat{\phi}_{wk} = \frac{N_{wk} + \eta}{N_k + W\eta} \qquad \hat{\theta}_{kj} = \frac{N_{kj} + \alpha}{N_j + K\alpha}. \qquad (15)$$

In the case of VB, the expected values of the posterior Dirichlets in (7) and (8) are used in prediction, leading to estimates for $\phi$ and $\theta$ of the same form as (15). However, for VB, an offset of $-0.5$ is found in update equation (10) while it is not found in the estimates used for prediction.

Table 1: Data sets used in experiments

| NAME | $D$ | $W$ | $N_{train}$ | $D_{test}$ |
|---|---|---|---|---|
| CRAN | 979 | 3,763 | 81,773 | 210 |
| KOS | 3,000 | 6,906 | 410,595 | 215 |
| MED | 9,300 | 5,995 | 886,306 | 169 |
| NIPS | 1,500 | 12,419 | 1,932,365 | 92 |
| NEWS | 19,500 | 27,059 | 2,057,207 | 249 |
| NYT | 6,800 | 16,253 | 3,768,969 | 139 |
| PAT | 6,500 | 19,447 | 14,328,094 | 106 |

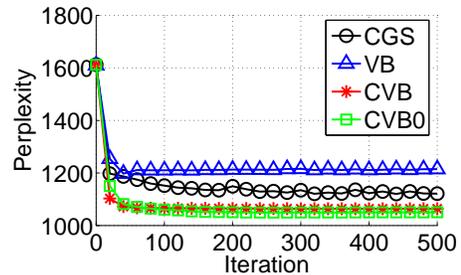

Figure 2: Convergence plot showing perplexities on MED, K=40; hyperparameters learned through Minka's update.

The knowledge that VB's update equation contains an effective offset of up to $-0.5$ suggests the use of an alternative estimate for prediction:

$$\hat{\phi}_{wk} \propto \frac{\exp(\psi(N_{wk} + \eta))}{\exp(\psi(N_k + W\eta))} \quad \hat{\theta}_{kj} \propto \frac{\exp(\psi(N_{kj} + \alpha))}{\exp(\psi(N_j + K\alpha))}. \quad (16)$$

Note the similarity that these estimates bear to the VB update (10). Essentially, the $-0.5$ offset is introduced into these estimates just as they are found in the update. Another way to mimic this behavior is to use $\alpha + 0.5$ and $\eta + 0.5$ during learning and then use $\alpha$ and $\eta$ for prediction, using the estimates in (15). We find that correcting this "mismatch" between the update and the estimate reduces the performance gap between VB and the other algorithms. Perhaps this phenomenon bears relationships to the observation of Wainwright [2006], who shows that it certain cases, it is beneficial to use the same algorithm for both learning and prediction, even if that algorithm is approximate rather than exact.

## 4 EXPERIMENTS

Seven different text data sets are used to evaluate the performance of these algorithms: Cranfield-subset (CRAN), Kos (KOS), Medline-subset (MED), NIPS (NIPS), 20 Newsgroups (NEWS), NYT-subset (NYT), and Patent (PAT). Several of these data sets are available online at the UCI ML Repository [Asuncion and Newman, 2007]. The characteristics of these data sets are summarized in Table 1.

Each data set is separated into a training set and a test set. We learn the model on the training set, and then we measure the performance of the algorithms on the test set. We also have a separate validation set of the same size as the test set that can be used to tune the hyperparameters. To evaluate accuracy, we use perplexity, a widely-used metric in the topic modeling community. While perplexity is a somewhat indirect measure of predictive performance, it is nonetheless a useful characterization of the predictive quality of a language model and has been shown to be well-correlated with other measures of performance such word-error rate in speech recognition [Klakow and Peters, 2002]. We also report precision/recall statistics.

We describe how perplexity is computed. For each of our algorithms, we perform runs lasting 500 iterations and we obtain the estimate $\hat{\phi}_{wk}$ at the end of each of those runs. To obtain $\hat{\theta}_{kj}$, one must learn the topic assignments on the first half of each document in the test set while holding $\hat{\phi}_{wk}$ fixed. For this fold-in procedure, we use the same learning algorithm that we used for training. Perplexity is evaluated on the second half of each document in the test set, given $\hat{\phi}_{wk}$ and $\hat{\theta}_{jk}$. For CGS, one can average over multiple samples (where S is the number of samples to average over):

$$\log p(\mathbf{x}^{\text{test}}) = \sum_{jw} N_{jw} \log \frac{1}{S} \sum_s \sum_k \hat{\theta}^s_{kj} \hat{\phi}^s_{wk} \ .$$

In our experiments we don't perform averaging over samples for CGS (other than in Figure 7 where we explicitly investigate averaging), both for computational reasons and to provide a fair comparison to the other algorithms. Using a single sample from CGS is consistent with its use as an efficient stochastic "mode-finder" to find a set of interpretable topics for a document set.

For each experiment, we perform three different runs using different initializations, and report the average of these perplexities. Usually these perplexities are similar to each other across different initializations (e.g. ± 10 or less).

### 4.1 PERPLEXITY RESULTS

In our first set of experiments, we investigate the effects of learning the hyperparameters during training using Minka's fixed point updates. We compare CGS, VB, CVB, and CVB0 in this set of experiments and leave out MAP since Minka's update does not apply to MAP. For each run, we initialize the hyperparameters to $\alpha = 0.5$, $\eta = 0.5$ and turn on Minka's updates after 15 iterations to prevent numerical instabilities. Every other iteration, we compute perplexity on the validation set to allow us to perform early-stopping if necessary. Figure 2 shows the test perplexity as a function of iteration for each algorithm on the MED data set. These perplexity results suggest that CVB and CVB0 outperform VB when Minka's updates are used. The reason is because the learned hyperparameters for VB are too small and do not correct for the effective $-0.5$ offset found in the VB update equations. Also, CGS converges more slowly than the deterministic algorithms.

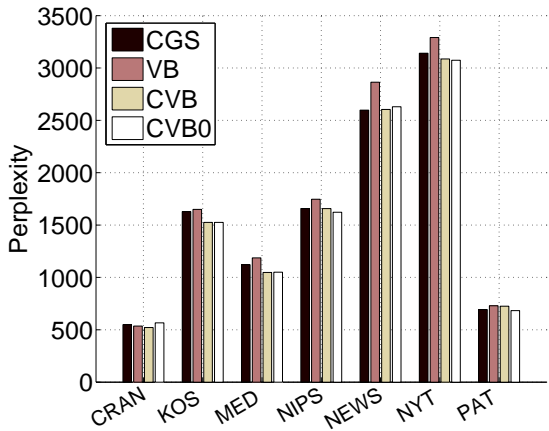

Figure 3: Perplexities achieved with hyperparameter learning through Minka's update, on various data sets, K=40.

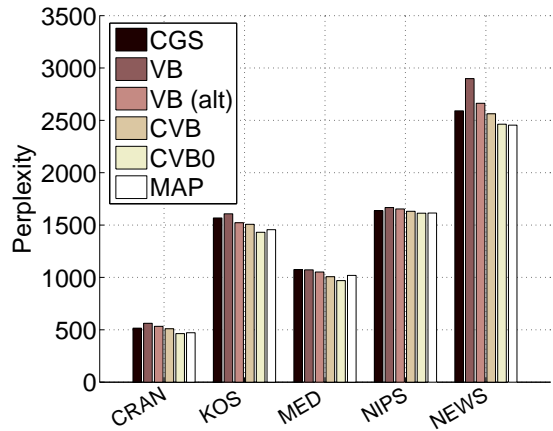

Figure 5: Perplexities achieved through grid search, K=40.

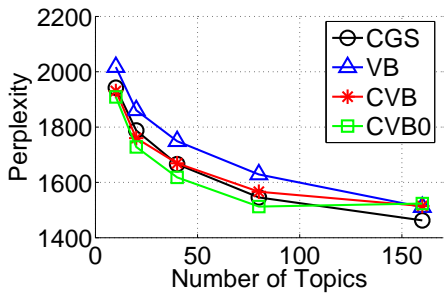

Figure 4: Perplexity as a function of number of topics, on NIPS, with Minka's update enabled.

In Figure 3, we show the final perplexities achieved with hyperparameter learning (through Minka's update), on each data set. VB performs worse on several of the data sets compared to the other algorithms. We also found that CVB0 usually learns the highest level of smoothing, followed by CVB, while Minka's updates for VB learn small values for $\alpha, \eta$.

In our experiments thus far, we fixed the number at topics at $K = 40$. In Figure 4, we vary the number of topics from 10 to 160. In this experiment, CGS/CVB/CVB0 perform similarly, while VB learns less accurate solutions. Interestingly, the CVB0 perplexity at $K = 160$ is higher than the perplexity at $K = 80$. This is due to the fact that a high value for $\eta$ was learned for CVB0. When we set $\eta = 0.13$ (to the $K = 80$ level), the CVB0 perplexity is 1464, matching CGS. These results suggest that learning hyperparameters during training (using Minka's updates) does not necessarily lead to the optimal solution in terms of test perplexity.

In the next set of experiments, we use a grid of hyperparameters for each of $\alpha$ and $\eta$, [0.01, 0.1, 0.25, 0.5, 0.75, 1], and we run the algorithms for each combination of hyperparameters. We include MAP in this set of experiments, and we shift the grid to the right by 1 for MAP (since hyperparameters less than 1 cause MAP to have negative probabilities). We perform grid search on the validation set, and we find the best hyperparameter settings (according to validation set perplexity) and use the corresponding estimates for prediction. For VB, we report both the standard perplexity calculation and the alternative calculation that was detailed previously in (16).

In Figure 5, we report the results achieved through performing grid search. The differences between VB (with the alternative calculation) and CVB have largely vanished. This is due to the fact that we are using larger values for the hyperparameters, which allows VB to reach parity with the other algorithms. The alternative prediction scheme for VB also helps to reduce the perplexity gap, especially for the NEWS data set. Interestingly, CVB0 appears to perform slightly better than the other algorithms.

Figure 6 shows the test perplexity of each method as a function of $\eta$. It is visually apparent that VB and MAP perform better when their hyperparameter values are offset by 0.5 and 1, respectively, relative to the other methods. While this picture is not as clear-cut for every data set (since the approximate VB update holds only when $n > 1$), we have consistently observed that the minimum perplexities achieved by VB are at hyperparameter values that are higher than the ones used by the other algorithms.

In the previous experiments, we used one sample for CGS to compute perplexity. With enough samples, CGS should be the most accurate algorithm. In Figure 7, we show the effects of averaging over 10 different samples for CGS, taken over 10 different runs, and find that CGS gains substantially from averaging samples. It is also possible for other methods like CVB0 to average over their local posterior "modes" but we found the resulting gain is not as great.

We also tested whether the algorithms would perform similarly in cases where the training set size is very small or the number of topics is very high. We ran VB and CVB with grid search on half of CRAN and achieved virtually the same perplexities. We also ran VB and CVB on CRAN with $K = 100$, and only found a 16-point perplexity gap.

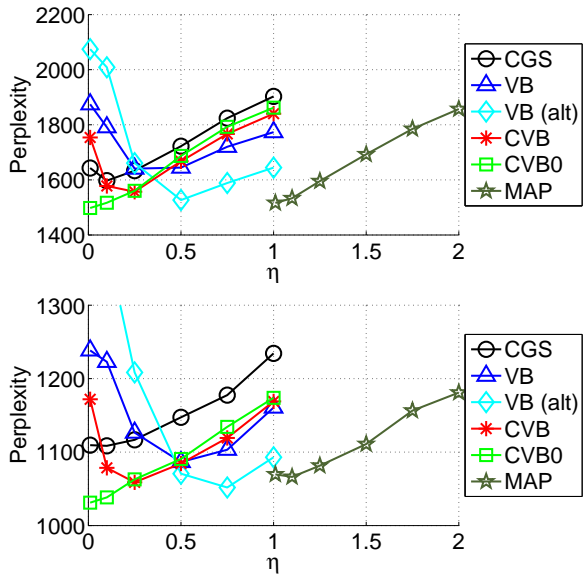

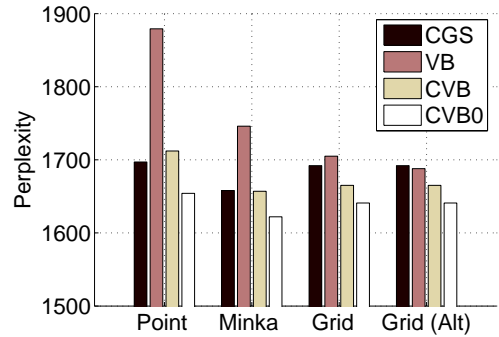

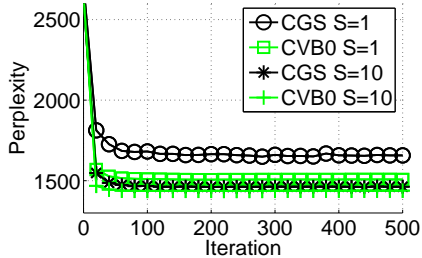

Figure 6: TOP: KOS, K=40. BOTTOM: MED, K=40. Perplexity as a function of $\eta$. We fixed $\alpha$ to 0.5 (1.5 for MAP). Relative to the other curves, VB and MAP curves are in effect shifted right by approximately 0.5 and 1.

Figure 7: The effect of averaging over 10 samples/modes on KOS, K=40.

Figure 8: POINT: When $\alpha = 0.1$, $\eta = 0.1$, VB performs substantially worse than other methods. MINKA: When using Minka's updates, the differences are less prominent. GRID: When grid search is performed, differences diminish even more, especially with the alternative estimates.

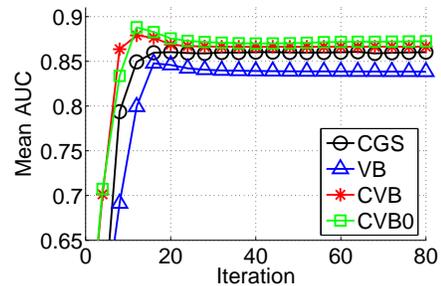

Figure 9: Mean AUC achieved on NEWS, K=40, with Minka's update.

To summarize our perplexity results, we juxtapose three different ways of setting hyperparameters in Figure 8, for NIPS, $K = 40$. The first way is to have the same arbitrary values used across all algorithms (e.g. $\alpha = 0.1$, $\eta = 0.1$). The second way is to learn the hyperparameters through Minka's update. The third way way is to find the hyperparameters by grid search. For the third way, we also show the VB perplexity achieved by the alternative estimates.

### 4.2 PRECISION/RECALL RESULTS

We also calculated precision/recall statistics on the NEWS data set. Since each document in NEWS is associated with one of twenty newsgroups, one can label each document by its corresponding newsgroup. It is possible to use the topic model for classification and to compute precision/recall statistics. In Figure 9, we show the mean area under the ROC curve (AUC) achieved by CGS, VB, CVB, and CVB0 with hyperparameter learning through Minka's update. We also performed grid search over $\alpha$, $\eta$ and found that each method was able to achieve similar statistics. For instance, on NEWS, $K = 10$, each algorithm achieved the same area under the ROC curve (0.90) and mean average precision (0.14). These results are consistent with the perplexity results in the previous section.

## 5　COMPUTATIONAL EFFICIENCY

While the algorithms can give similarly accurate solutions, some of these algorithms are more efficient than others. VB contains digamma functions which are computationally expensive, while CVB requires the maintenance of variance counts. Meanwhile, the stochastic nature of CGS causes it to converge more slowly than the deterministic algorithms.

In practice, we advocate using CVB0 since 1) it is faster than VB/CVB given that there are no calls to digamma or variance counts to maintain; 2) it converges more quickly than CGS since it is deterministic; 3) it does not have MAP's $-1$ offset issue. Furthermore, our empirical results suggest that CVB0 learns models that are as good or better (predictively) than those learned by the other algorithms.

These algorithms can be parallelized over multiple processors as well. The updates in MAP estimation can be performed in parallel without affecting the fixed point since MAP is an EM algorithm [Neal and Hinton, 1998]. Since the other algorithms are very closely related to MAP there is confidence that performing parallel updates over tokens for the other algorithms would lead to good results as well.

Table 2: Timing results (in seconds)

|  | MED | KOS | NIPS |
|---|---|---|---|
| VB | 151.6 | 73.8 | 126.0 |
| CVB | 25.1 | 9.0 | 21.7 |
| CGS | 18.2 | 3.8 | 10.0 |
| CVB0 | 9.5 | 4.0 | 8.4 |
| Parallel-CVB0 | 2.4 | 1.5 | 3.0 |

While non-collapsed algorithms such as MAP and VB can be readily parallelized, the collapsed algorithms are sequential, and thus there has not been a theoretical basis for parallelizing CVB or CGS (although good empirical results have been achieved for approximate parallel CGS [Newman et al., 2008]). We expect that a version of CVB0 that parallelizes over tokens would converge to the same quality of solution as sequential CVB0, since CVB0 is essentially MAP but without the $-1$ offset[4].

In Table 2, we show timing results for VB, CVB, CGS, and CVB0 on MED, KOS, and NIPS, with $K = 10$. We record the amount of time it takes for each algorithm to pass a fixed perplexity threshold (the same for each algorithm). Since VB contains many calls to the digamma function, it is slower than the other algorithms. Meanwhile, CGS needs more iterations before it can reach the same perplexity, since it is stochastic. We see that CVB0 is computationally the fastest approach among these algorithms. We also parallelized CVB0 on a machine with 8 cores and find that a topic model with coherent topics can be learned in 1.5 seconds for KOS. These results suggest that it is feasible to learn topic models in near real-time for small corpora.

## 6 RELATED WORK & CONCLUSIONS

Some of these algorithms have been compared to each other in previous work. Teh et al. [2007] formulate the CVB algorithm and empirically compare it to VB, while Mukherjee and Blei [2009] theoretically analyze the differences between VB and CVB and give cases for when CVB should perform better than VB. Welling et al. [2008b] also compare the algorithms and introduce a hybrid CGS/VB algorithm. In all these studies, low values of $\eta$ and $\alpha$ were used for each algorithm, including VB. Our insights suggest that VB requires more smoothing in order to match the performance of the other algorithms.

The similarities between PLSA and LDA have been noted in the past [Girolami and Kaban, 2003]. Others have unified similar deterministic latent variable models [Welling et al., 2008a] and matrix factorization techniques [Singh and Gordon, 2008]. In this work, we highlight the similarities between various learning algorithms for LDA.

While we focused on LDA and PLSA in this paper, we believe that the insights gained are relevant to learning in general directed graphical models with Dirichlet priors, and generalizing these results to other models is an interesting avenue to pursue in the future.

In conclusion, we have found that the update equations for these algorithms are closely connected, and that using the appropriate hyperparameters causes the performance differences between these algorithms to largely disappear. These insights suggest that hyperparameters play a large role in learning accurate topic models. Our comparative study also showed that there exist accurate and efficient learning algorithms for LDA and that these algorithms can be parallelized, allowing us to learn accurate models over thousands of documents in a matter of seconds.

### Acknowledgements

This work is supported in part by NSF Awards IIS-0083489 (PS, AA), IIS-0447903 and IIS-0535278 (MW), and an NSF graduate fellowship (AA), as well as ONR grants 00014-06-1-073 (MW) and N00014-08-1-1015 (PS). PS is also supported by a Google Research Award. YWT is supported by the Gatsby Charitable Foundation.

### References

A. Asuncion and D. Newman. UCI machine learning repository, 2007. URL http://www.ics.uci.edu/~mlearn/MLRepository.html.

H. Attias. A variational Bayesian framework for graphical models. In *NIPS 12*, pages 209–215. MIT Press, 2000.

D. M. Blei, A. Y. Ng, and M. I. Jordan. Latent Dirichlet allocation. *JMLR*, 3:993–1022, 2003.

W. Buntine and A. Jakulin. Discrete component analysis. *Lecture Notes in Computer Science*, 3940:1, 2006.

J.-T. Chien and M.-S. Wu. Adaptive Bayesian latent semantic analysis. *Audio, Speech, and Language Processing, IEEE Transactions on*, 16(1):198–207, 2008.

N. de Freitas and K. Barnard. Bayesian latent semantic analysis of multimedia databases. Technical Report TR-2001-15, University of British Columbia, 2001.

S. Deerwester, S. Dumais, G. Furnas, T. Landauer, and R. Harshman. Indexing by latent semantic analysis. *JASIS*, 41(6):391–407, 1990.

Z. Ghahramani and M. Beal. Variational inference for Bayesian mixtures of factor analysers. In *NIPS 12*, pages 449–455. MIT Press, 2000.

M. Girolami and A. Kaban. On an equivalence between PLSI and LDA. In *SIGIR '03*, pages 433–434. ACM New York, NY, USA, 2003.

T. L. Griffiths and M. Steyvers. Finding scientific topics. *PNAS*, 101(Suppl 1):5228–5235, 2004.

T. Hofmann. Unsupervised learning by probabilistic latent semantic analysis. *Machine Learning*, 42(1):177–196, 2001.

D. Klakow and J. Peters. Testing the correlation of word error rate and perplexity. *Speech Communication*, 38(1-2):19–28, 2002.

T. Minka. Estimating a Dirichlet distribution. 2000. URL http://research.microsoft.com/~minka/papers/dirichlet/.

I. Mukherjee and D. M. Blei. Relative performance guarantees for approximate inference in latent Dirichlet allocation. In *NIPS 21*, pages 1129–1136, 2009.

R. Neal and G. Hinton. A view of the EM algorithm that justifies incremental, sparse, and other variants. *Learning in graphical models*, 89:355–368, 1998.

D. Newman, A. Asuncion, P. Smyth, and M. Welling. Distributed inference for latent Dirichlet allocation. In *NIPS 20*, pages 1081–1088. MIT Press, 2008.

A. Singh and G. Gordon. A Unified View of Matrix Factorization Models. In *ECML PKDD*, pages 358–373. Springer, 2008.

Y. W. Teh, M. I. Jordan, M. J. Beal, and D. M. Blei. Hierarchical Dirichlet processes. *Journal of the American Statistical Association*, 101(476):1566–1581, 2006.

Y. W. Teh, D. Newman, and M. Welling. A collapsed variational Bayesian inference algorithm for latent Dirichlet allocation. In *NIPS 19*, pages 1353–1360. 2007.

M. J. Wainwright. Estimating the "wrong" graphical model: Benefits in the computation-limited setting. *JMLR*, 7:1829–1859, 2006.

H. M. Wallach. *Structured Topic Models for Language*. PhD thesis, University of Cambridge, 2008.

B. Wang and D. Titterington. Convergence and asymptotic normality of variational Bayesian approximations for exponential family models with missing values. In *UAI*, pages 577–584, 2004.

M. Welling, C. Chemudugunta, and N. Sutter. Deterministic latent variable models and their pitfalls. In *SIAM International Conference on Data Mining*, 2008a.

M. Welling, Y. W. Teh, and B. Kappen. Hybrid variational/MCMC inference in Bayesian networks. In *UAI*, volume 24, 2008b.

---

[4] If one wants convergence guarantees, one should also not remove the current token $ij$.